\documentclass[conference]{IEEEtran}
\IEEEoverridecommandlockouts
\usepackage{cite}
\usepackage{amsmath,amssymb,amsfonts}
\usepackage{algorithmic}
\usepackage{graphicx}
\usepackage{textcomp}
\usepackage{xcolor}
\usepackage{caption}
\usepackage{booktabs}
\usepackage{tabularx}
\usepackage{makecell}
\usepackage{multirow}
\usepackage{url}

\def\BibTeX{{\rm B\kern-.05em{\sc i\kern-.025em b}\kern-.08em
    T\kern-.1667em\lower.7ex\hbox{E}\kern-.125emX}}
\begin{document}

\title{Reinforced Model Merging}

\author{\IEEEauthorblockN{Jiaqi	Han$^1$, 
                          Jingwen Ye$^2$, 
                          Shunyu Liu$^3$, 
                          Haofei Zhang$^1$, 
                          Jie Song$^4$, 
                          Zunlei Feng$^4$, 
                          Mingli Song$^{1,\dag}$}\thanks{$^\dag$Corresponding author.}
\IEEEauthorblockA{
$^1$College of Computer Science and Technology, Zhejiang University, Hangzhou, China\\
$^2$National University of Singapore, Singapore\\
$^3$Nanyang Technological University, Singapore\\
$^4$School of Software Technology, Zhejiang University, Ningbo, China\\
\{jiaqihan, haofeizhang, sjie, zunleifeng, brooksong\}@zju.edu.cn, jingweny@nus.edu.sg, shunyu.liu@ntu.edu.sg}
}

\maketitle

\begin{abstract}
The success of large language models has garnered widespread attention for model merging techniques, especially training-free methods which combine model capabilities within the parameter space. However, two challenges remain: (1) uniform treatment of all parameters leads to performance degradation; (2) search-based algorithms are often inefficient. In this paper, we present an innovative framework termed Reinforced Model Merging (RMM), which encompasses an environment and agent tailored for merging tasks. These components interact to execute layer-wise merging actions, aiming to search the optimal merging architecture. Notably, RMM operates without any gradient computations on the original models, rendering it feasible for edge devices. Furthermore, by utilizing data subsets during the evaluation process, we addressed the bottleneck in the reward feedback phase, thereby accelerating RMM by up to 100 times. Extensive experiments demonstrate that RMM achieves state-of-the-art performance across various vision and NLP datasets and effectively overcomes the limitations of the existing baseline methods. Our code is available at \url{https://github.com/WuDiHJQ/Reinforced-Model-Merging}.

\end{abstract}

\section{Introduction}
In recent years, the rapid advances of deep learning have led to the emergence of an increasing number of pre-trained models~\cite{raffel2020exploring,dosovitskiy2020image,radford2021learning,touvron2023llama}, the majority of which have been released on open-source platforms for training purposes (e.g. \textit{Hugging Face}). The general knowledge contained in these models greatly facilitates the training for downstream tasks and profoundly transforms the manner of model acquisition. Furthermore, the pre-training and fine-tuning paradigm enables models to rapidly optimize for specific tasks, thereby improving the deployment efficiency of deep learning.

However, when attempting to extend the capabilities of existing models, several challenges inevitably arise, such as catastrophic forgetting~\cite{li2017learning} and excessive time overhead. Fortunately, numerous model reuse approaches~\cite{hinton2015distilling,ji2021show,he2022knowledge,fang2022up} have emerged to leverage these task-specific models. Among these, model merging constitutes a significant branch which combines multiple different models into a single versatile model. Early merging methods primarily concentrated in knowledge amalgamation (KA)~\cite{shen2019amalgamating,xie2022federated,zhang2022knowledge}, which involved generating compact features through specific training processes. Recently, training-free merging methods~\cite{wortsman2022model,ilharco2022editing,ainsworth2022git,yadav2023ties,yu2024language} have attracted considerable attention from researchers, which solely merge models in the parameter space without performing any gradient computations on the original models, hence substantially enhancing merging efficiency.

\begin{figure}[t]
    \centering
    \includegraphics[width=\columnwidth]{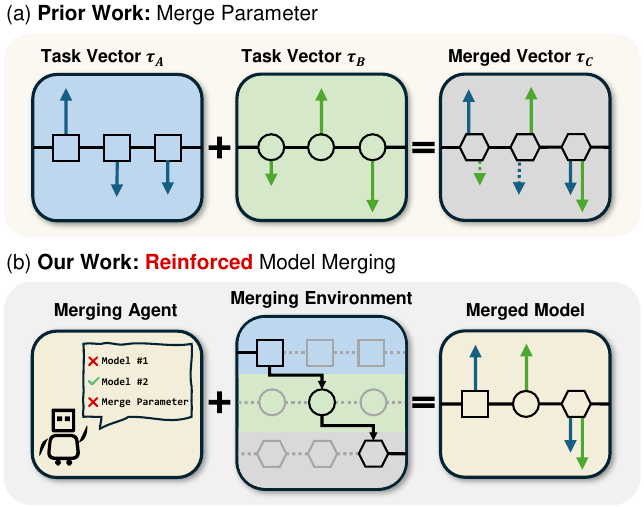}
    \caption{\textbf{Compared with prior methods.} (a) Prior training-free merging methods trim and align the task vector (i.e. the difference in parameter values between fine-tuned and pre-trained models) in parameter space. (b) RMM embeds RL into the merging framework, resulting in an automatic, search-based merging paradigm.} 
    \label{compare}
\end{figure}

Unfortunately, when merging models with vast domain gaps, current training-free methods often lead to substantial performance degradation in certain tasks. This primarily arises from the fact that models trained on different tasks acquire heterogeneous knowledge, rendering it impractical to apply a uniform approach to merge all parameters. Recently proposed search-based methods~\cite{akiba2024evolutionary} can alleviate this issue, but they highly rely on iterating and evaluating merged model on the whole dataset, which undoubtedly consumes a substantial amount of time. Moreover, searching for the merged models in a fixed mode is quite rigid and incapable of accommodating diverse scenarios and tasks.

To address the above issues, in this paper, we propose a flexible and effective framework termed Reinforced Model Merging (RMM). As shown in Fig.~\ref{compare}, RMM can model the merging task as an agent's decision-making process, enabling the merged model to be optimized layer by layer in a extensible environment automatically. In each episode, the agent will receive rewards evaluated by the merged model. To expedite this stage, we introduce the Dynamic Average Reward (DAR) mechanism, which only employs a small subset of data for evaluation and mitigates the instability in results due to the imbalanced distribution of evaluation data. We validate our approach across a variety of vision and NLP tasks, achieving performance improvements of 5.57\% and 8.56\%, respectively. These findings suggest that RMM not only reduces computational time but also enhances merging performance, thereby making its practical application viable. 

In summary, the key contributions of this work are:
\begin{itemize}
  \item  \textbf{RL-Driven Model Merging}: We present the first RL-based merging approach, which avoids complex gradient computations and optimizes merged architectures within an adjustable environment, building the powerful models with minimal cost.
  \item  \textbf{Up to 100× Faster RMM}: Our RMM agent employs DAR mechanism, requiring merely a small fraction of data for evaluation in each search episode. This approach can achieve near-full dataset performance while substantially enhancing merging efficiency.  
  \item  \textbf{State-of-the-Art Performance}: We demonstrate the effectiveness of RMM across various benchmark vision and language datasets, showing that it can identify better merging architectures and improve existing algorithms.
\end{itemize}

\begin{figure*}[t]
  \centering
  \includegraphics[width=\linewidth]{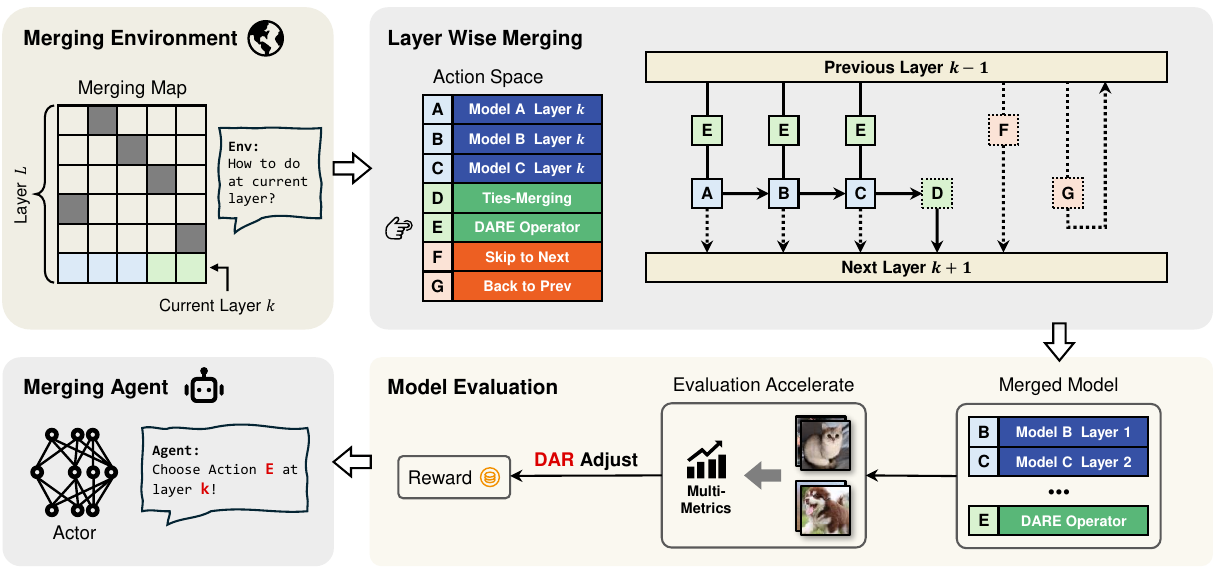}
  \caption{
  \textbf{Overview of the proposed RMM.} Our framework incorporates RL into the merging procedure, searching the layer-wise optimal architecture through the interactions between the environment and agent. In each step, the merging map is presented as state to the agent and prompts it to make wise decisions. At the end of an episode, the merged model will be handed over to the environment for evaluation and return a reward to optimize the agent's decisions. Repeatedly iterate until convergence.
  }
  \label{RMMoverview}
\end{figure*}

\section{Methodology}

\subsection{Preliminaries}
Let the set $\{f_1,f_2,...,f_N\}$ represent $N$ models prepared for merging, each fine-tuned from the pre-trained model $f_{pt}$ using their respective datasets $\{X_i, Y_i\}_{i=1}^N$. Among them, $f_i^l$ is the $l$-th $(l\in\{1,2,...,L\})$ layer in $i$-th model $f_i$, where $L$ denotes the number of layers of the fine-tuned network.

In this work, our goal is to propose a flexible and extensible framework that can automatically handle the layer-wise merging procedure. While current model merging methods can improve performance across diverse tasks and even develop capabilities for new tasks, they often struggle to identify the optimal solution for specific scenarios. To address this, we introduce RMM for model merging, where the agent navigates through model layers in the designed environment, executes specific merging actions, and receives corresponding rewards. Fig.~\ref{RMMoverview} provides an overview of our framework.

\subsection{Merging Agent}
In RMM, we propose the merging agent to browse multiple models layer by layer to connect networks and create merged models. The proposed merging agent receives the state $s_t$ from the environment at step $t$ as input, makes policy within the tailored action space, and executes merging operations based on the decision results. Assuming the agent is dealing with the $k$-th layer, the merging step can be represented as:
\begin{equation}
    a^*_t = {\arg \max}_{a \in A}Q_{\pi}(s_t,a)
\end{equation}
\begin{equation}
    F^*_t= \Psi(f_1^k,f_2^k,...,f_N^k,a^*_t)
\end{equation}
where $a^*_t$ denotes the action which yields the maximum $Q(s,a)$ value according to the policy $\pi$, and the function $\Psi()$ generates the optimal merged layer $F^*_t$ based on the layer parameters and determined action.

It is evident that the effective action space $A$ and function $\Psi()$ plays a pivotal role in obtaining a optimal merged model. To this end, we design a set of extensible merging actions for the RL-based method, primarily comprising three types of actions: model action, merging action, and layer action. Next, we will introduce each of them separately.

\textbf{Model Action:} Model actions $A_{Model}$ are mainly focus on preserving the parameters of the basic models. RMM allows the integration of task-specific knowledge into the merged model, thereby enhancing its capabilities on those tasks. This setting can effectively alleviate the performance degradation on individual tasks caused by existing merging methods, as they uniformly and equally handle all model parameters. The model actions can be executed as:
\begin{equation}
    F^*_t= \Psi(f_1^k,f_2^k,...,f_N^k,a^*_t) = f_{a^*_t}^k, a^*_t \in A_{Model}
\end{equation}

\textbf{Merging Action:} However, relying solely on combinations within the original knowledge space is inadequate. We introduce a set of merging actions $A_{Merge}$ which are composed of $M$ model merging operators (e.g. Task Arithmetic~\cite{ilharco2022editing}, Ties-Merging~\cite{yadav2023ties}), denote as $\{O_1,O_2,...,O_M\}$. These operators have demonstrated efficacy in aggregating knowledge across various tasks, and are crucial in establishing connections between disparate layers of knowledge within the merged model. The merge actions operated as follows:
\begin{equation}
\begin{split}
    F^*_t &= \Psi(f_1^k,f_2^k,...,f_N^k,a^*_t) \\
          &= O_{a^*_t}(f_1^k,f_2^k,...,f_N^k), a^*_t \in A_{Merge}
\end{split}
\end{equation}

\textbf{Layer Action:} The two types of actions mentioned above will automatically advance to the next layer for decision-making after acting on a specific position. Inspired by the work~\cite{akiba2024evolutionary}, we propose two layer actions $A_{Layer}$ called skipping and backing. Specifically, we allow the agent to neglect the current layer and directly enter the next layer for jump connections when making decisions. Alternatively, the agent can also return to the previous layer in the model to achieve hierarchical stacking. Notably, both layer actions solely modify the value of $k$, and the optimal return layer $F^*_t$ of function $\Psi()$ is None.

\subsection{Merging Environment}
To align with the aforementioned action space, we present the merging environment that empowers the agent to explore optimal merging architectures, which are typically challenging for humans to identify. Specifically, we design a matrix merging map, denoted as $s_t\in\mathbb{R}^{(M+N) \times L}$, as the state information at step $t$, which serves as the input for agent to determine the most valuable action at current position. The first dimension of $s_t$ reflects the decision regarding $A_{Model}$ and $A_{Merge}$, whereas the second dimension delineates the merging position of the current step. The initial state $s_0$, is set to a zero matrix.

To ensure that the framework can be treated as a Markov Decision Process (MDP) and optimized by RL algorithms, we track and record the trajectory of the agent's decision within the merging map $s_t$. At step $t$, when the agent opts for action $a_t$ during the decision-making process at layer $k$, the corresponding state $s_t$ will be updated as:
\begin{equation}
\label{eqn:update}
s_{t+1}=
\begin{cases}
s_{t}(i,j)+1   &\text{if } i=a_t \text{ and } j=k \\
s_{t}(i,j)     &\text{otherwise}
\end{cases}
\end{equation}

According to Eqn.~\ref{eqn:update}, the agent's action history will be retained until the $T$ step generation of a complete trajectory. Subsequently, we can assemble the optimal layer $F^*_t$ at each step to construct final merged model and calculate the reward derived from the policy $\pi$ as follows:
\begin{equation}
    F_M = F^*_1 \circ F^*_2 \circ ... \circ F^*_T
\end{equation}
\begin{equation}
    R_{\pi} = \frac{1}{N}\sum\limits_{i=1}^{N}R_{metric}(F_M(X_i), Y_i)
\end{equation}
where the symbol $\circ$ denotes the operator that connects model layers, $R_{metric}$ is employed to evaluate the performance of the merged model $F_M$ on various datasets $\{X_i, Y_i\}$.

\begin{table*}[t]
  \centering
  \caption{\textbf{Vision task results on ViT-S/32 and ViT-B/16.} We report the results of each merging methods on various datasets.}
  \begin{tabularx}{\linewidth}{p{2.5cm}
        X<{\centering}X<{\centering}X<{\centering}X<{\centering}X<{\centering}
        X<{\centering}X<{\centering}X<{\centering}X<{\centering}X<{\centering}}
    \toprule 
        \textbf{Model ($\rightarrow$)}  
        &\multicolumn{5}{c}{\textbf{ViT-S/32}}
        &\multicolumn{5}{c}{\textbf{ViT-B/16}}\\
        \cmidrule(lr){1-1}\cmidrule(lr){2-6} \cmidrule(lr){7-11}
        \textbf{Method ($\downarrow$)}  
                         &\textbf{CIFAR10} 
                         &\textbf{CIFAR100}
                         &\textbf{CUB}
                         &\textbf{Dogs} 
                         &\textbf{Average}
                         &\textbf{CIFAR10} 
                         &\textbf{CIFAR100}
                         &\textbf{CUB}
                         &\textbf{Dogs} 
                         &\textbf{Average}\\
    \midrule
        Fine-Tuned            
        & 99.31  & 92.24  & 70.29  & 77.75  & 84.90  & 99.65  & 95.25  & 86.40 & 87.55 & 92.21 \\
        Multi-Task
        & 99.38  & 92.25  & 73.10  & 78.45  & 85.79  & 99.70  & 95.30  & 85.98 & 87.45 & 92.10 \\
    \midrule
        Weight Averaging
        & 97.34  & 83.34  & 51.24  & 68.60  & 75.13  & 98.53  & 81.35  & 59.28 & 76.61 & 78.94 \\
        Task Arithmetic
        & 97.64  & 83.82  & 51.36  & 68.74  & 75.39  & 98.62  & 81.73  & 59.37 & 76.62 & 79.08 \\
        Ties-Merging
        & 97.34  & 82.93  & 51.51  & 68.61  & 75.09  & 98.48  & 80.62  & 60.34 & 76.84 & 79.07 \\
        DARE
        & 97.64  & 83.79  & 51.46  & 68.85  & 75.43  & 98.61  & 81.76  & 59.25 & 76.78 & 79.10 \\
    \midrule
        RMM (Ours)
        & \textbf{98.66}  & \textbf{87.72}  & \textbf{59.46}  & \textbf{72.31}  & \textbf{79.53}  & \textbf{99.41}  & \textbf{89.53}  & \textbf{73.62} & \textbf{81.86} & \textbf{86.10} \\
    \bottomrule
  \end{tabularx}
  \label{vit}
\end{table*}
\begin{table}[t]
  \centering
  \caption{\textbf{Fine-grained Merging results on ViT-B/16.} We report the performance of models merged from CUB-200 and Stanford Dogs across various methods.}
  \begin{tabularx}{\linewidth}{p{2.2cm}X<{\centering}X<{\centering}X<{\centering}}
    \toprule
        \textbf{Method}  
                         &\textbf{CUB} 
                         &\textbf{Dogs}
                         &\textbf{Average}\\
    \midrule
        Fine-Tuned                    
        & 84.35 & 86.06 & 85.20   \\
        Multi-Task
        & 84.31 & 85.39 & 84.85    \\
    \midrule
        Weight Averaging
        & 46.20 & 70.35 & 58.27    \\
        Task Arithmetic
        & 46.58 & 70.48 & 58.53    \\
        Ties-Merging
        & 47.70 & 70.79 & 59.24    \\
        DARE
        & 46.44 & 70.43 & 58.43    \\
    \midrule
        RMM (Ours)
        & \textbf{69.02} & \textbf{78.58} & \textbf{73.80}    \\
    \bottomrule
  \end{tabularx}
  \label{cub_dogs}
\end{table}

\subsection{RL Optimization}
In this section, we will detail the optimization process of RMM, which is founded on the proposed merging agent and environment. First, our reward feedback $R_{\pi}$ is composed of the average metrics from the merged model across multiple tasks, and this signal is clearly non-differentiable. To address this, we adopt a policy gradient method called Proximal Policy Optimization (PPO)~\cite{schulman2017proximal} to optimize the decision-making of the agent. The merging agent consists of two components: the actor, which predicts the probabilities $\pi_\theta(a_t|s_t)$ of taking various actions in the current state $s_t$, and the critic, which evaluates the expected reward $V_\pi(s_t)$ obtained in that state. PPO is an off-policy algorithm in which the agent $\pi_\theta'$ interacting with the environment is not the same as the agent $\pi_\theta$ updating the parameters. By employing the technique of importance sampling, PPO can sample a small batch of trajectory to optimize the parameters multiple times. The final optimization goal can be summarized as follows:
\begin{equation}
\begin{split}
    J(\theta) = \sum\limits_{(s_t,a_t)}min(\frac{\pi_\theta(a_t|s_t)}{\pi_{\theta'}(a_t|s_t)}A_{\pi'}(s_t,a_t), \\
     clip(\frac{\pi_\theta(a_t|s_t)}{\pi_{\theta'}(a_t|s_t)},1-\varepsilon,1+\varepsilon)A_{\pi'}(s_t,a_t))
\end{split}
\label{eqn:ppo}
\end{equation}
where $\frac{\pi_\theta(a_t|s_t)}{\pi_{\theta'}(a_t|s_t)}$ is the probability ratio term used for importance sampling, $clip(\frac{\pi_\theta(a_t|s_t)}{\pi_{\theta'}(a_t|s_t)},1-\varepsilon,1+\varepsilon)$ adjusts the probability ratio to ensure that it will not exceed the interval $[1-\varepsilon,1+\varepsilon]$, $A_{\pi'}(s_t,a_t)$ represents advantage function, which is the difference between the reward obtained from taking action $a_t$ in state $s_t$ and the baseline in current state. Here, we set $V_\pi(s_t)$ as baseline, then $A_{\pi'}(s_t,a_t)$ can be obtained as:
\begin{equation}
    A_{\pi'}(s_t,a_t) = R_{\pi'} - V_{\pi'}(s_t)
\label{eqn:ppo}
\end{equation}

The reward distribution in RMM is relatively sparse, as it requires $T$ steps of merging in each episode to achieve model evaluation. Consequently, adopting PPO based on the Actor-Critic framework can yield more stable performance. It is worth noting that RMM optimizes the merging model through dynamic interaction between the agent and the environment, so when the sophistication of the action space escalates (i.e. $N$ and $M$ become larger), the complexity of the architecture search procedure will not increase significantly.

\subsection{Merging Accelerate}

During the optimization phase of RMM, the agent's policy is updated based on the evaluation reward from the merged model. Thus, conducting evaluations on test data for each task is inevitable. Although our method do without calculate a large number of gradients on the original model parameters, evaluation still incurs a significant time overhead.

To address this, we propose a merging accelerate method called Dynamic Average Reward (DAR). DAR can achieve up to 100× acceleration in the interaction process while ensuring minimal impact on the performance of the merged model. In other words, by only using small fraction of data for evaluation, the agent can obtain an estimation of the current architecture's metrics. During the initial phase of RMM, the estimation metrics are sufficient to guide the agent in selecting roughly correct merging actions. But once the agent's policy stabilizes, metrics based on small batches may exert a detrimental impact. Therefore, we introduce DAR to modulate rewards at different phases of RMM. In the early stage, the agent pays more attention to the rewards derived from limited data within the current architecture. In contrast, during the later stages, the agent needs to consider the overall evaluation results to achieve a comprehensive assessment of analogous architectures. The equation of DAR is as follows:
\begin{equation}
    R_{t} = \lambda \frac{t}{T_{max}} R_{t-1} + (1 - \lambda \frac{t}{T_{max}})R_{t} 
\label{eqn:DAR}
\end{equation}
where $R_{t}$ and $R_{t-1}$ represent the rewards of the current and previous episode, $T_{max}$ is the maximum number of steps in the whole RMM, and $\lambda$ is the scaling coefficient that allows us to dynamically adjust the importance of each reward. It should be noted that we do not propagate the previous rewards throughout the entire RMM process, instead, $\lambda$ will be set to 0 when the dataloader is cleared.

\section{Experiments}

In this section, we conducted extensive experiments on vision and NLP tasks to thoroughly validate the superiority of RMM. And the results indicate that RMM can achieve exceptional performance across various merging settings, significantly outperforming its state-of-the-art competitors. More experimental details can be found in supplementary material.

\subsection{Baseline Methods.} 
The proposed RMM can theoretically extend to any task arithmetic algorithm. Consequently, we compare it with four recent baseline methods: Weight Averaging~\cite{wortsman2022model}, Task Arithmetic~\cite{ilharco2022editing}, Ties-Merging~\cite{yadav2023ties}, and DARE~\cite{yu2024language}. We demonstrate the performance of RMM incorporating these algorithms as merging actions. For a fair comparison, we utilize the same hyper-parameter settings in merging. Besides above-mentioned, we also present the performance of fine-tuned models and multi-task models which trained from all datasets.

\subsection{Merging Vision Models}
\subsubsection{Datasets and Models} 
For vision models, our experiments are primarily conducted on four mainstream datasets for image classification tasks containing CIFAR-10~\cite{krizhevsky2009learning}, CIFAR-100~\cite{krizhevsky2009learning}, CUB-200~\cite{wah2011caltech}, and Stanford Dogs~\cite{dataset2011novel}. Each dataset was divided into two non-overlapping subsets with an equal number of categories, which were then used to fine-tune four pairs of models. Subsequently, we merge them in pairs using different methods and report the average accuracy of each group and the overall performance. In this experiment, we adopted the widely used ViT family~\cite{dosovitskiy2020image} as the model architecture, including ViT-S/32 and ViT-B/16.

\begin{table}[t]
  \centering
  \caption{\textbf{NLP task results on T5-Small and T5-Base.} We conducted extensive experiments on seven language datasets, dividing them into two merging tasks to separately report the performance of various merging methods.}
  \begin{tabularx}{\linewidth}{p{2.2cm}X<{\centering}X<{\centering}X<{\centering}X<{\centering}}
    \toprule
        \textbf{Task ($\rightarrow$)}  
        &\multicolumn{2}{c}{\textbf{QA Tasks}}
        &\multicolumn{2}{c}{\textbf{Mixed NLP Tasks}}\\
        \cmidrule(lr){1-1}\cmidrule(lr){2-3} \cmidrule(lr){4-5}
        \textbf{Method ($\downarrow$)} 
                         &\textbf{T5-Small} 
                         &\textbf{T5-Base}   
                         &\textbf{T5-Small} 
                         &\textbf{T5-Base}   \\
    \midrule
        Fine-Tuned                     
        & 88.36  & 92.72  & 69.02  & 75.43  \\
        Multi-Task
        & 89.19  & 92.71  & 68.72  & 76.75  \\
    \midrule
        Weight Averaging        
        & 69.21  & 85.55  & 49.59  & 56.28  \\
        Task Arithmetic
        & 69.77  & 86.34  & 50.86  & 56.59  \\
        Ties-Merging
        & 67.05  & 85.42  & 51.27  & 56.54  \\
        DARE
        & 69.89  & 86.62  & 51.02  & 57.01  \\
    \midrule
        RMM (Ours)
        & \textbf{80.52}  & \textbf{88.94}  & \textbf{60.59}  & \textbf{68.99}  \\
    \bottomrule
  \end{tabularx}
  \label{t5}
\end{table}

\subsubsection{Results and Analysis}
Table~\ref{vit} summarizes the results of ViT-S/32 and ViT-B/16 across multiple dataset pairs respectively. In image classification task, merging agent receives rewards based solely on 10\% of the training data in each episode. And we utilize top-1 accuracy as the evaluation metric to assess each merged model on the validation dataset. Compared to traditional merging methods, RMM achieves average improvements of 4.14\% and 7.00\% for ViT-S/32 and ViT-B/16. Specifically, for some basic vision tasks such as CIFAR-10 and CIFAR-100, RMM outperforms the state-of-the-art method by 0.90\% to 5.83\%. Furthermore, when addressing more challenging fine-grained merging tasks such as CUB-200 and Stanford Dogs, RMM demonstrates a substantial performance boost of up to 13.28\%, highlighting its significant potential for effectively exploring multi-model knowledge merging.

\subsubsection{Fine-grained Merging}
To further study the effectiveness of RMM, we conduct experiments on more challenging fine-grained cross-domain tasks. Specifically, we fine-tune ViT-B/16 on the complete CUB-200 and Stanford Dogs datasets, and subsequently merge them with various methods. It is worth noting that these datasets correspond to two entirely distinct species classification tasks, indicating that the knowledge learned by the models is likely to differ significantly.

Due to the obstacle caused by the domain gap, resolving sign conflict and value redundancy at the parameter level will inevitably result in a merged model with limited capabilities. Such merging is constrained, as it fails to consider the knowledge adaptation across various hierarchical levels of the models. As shown in Table~\ref{cub_dogs}, previous merging algorithms yield a merged model with poor generalization ability on CUB-200, with nearly half of the performance compared to fine-tuned model. In contrast, RMM exhibit significant advancements in two datasets, up to 21.32\% and 7.79\%, respectively. This suggests that the proposed method overcomes the limitations of traditional merging approaches as well as elevating the capabilities of model merging to a new level.

\subsection{Merging NLP Models}
\subsubsection{Datasets and Models} 
For NLP tasks, our experiments are mainly conducted on seven diverse datasets from four major tasks, including Question Answering (QASC~\cite{khot2020qasc}, WikiQA~\cite{yang2015wikiqa}, QuaRTz~\cite{tafjord2019quartz}), Paraphrase Identification (PAWS~\cite{zhang2019paws}), Sentence Completion (Story Cloze~\cite{mostafazadeh2017lsdsem}) and Coreference Resolution (Winogrande~\cite{sakaguchi2021winogrande} and WSC~\cite{levesque2012winograd}). We categorized the above data into two merging tasks: the first task involves three Question Answering datasets, while the second task merges four remaining datasets which regard to distinct scenarios. In this experiment, we employed the advanced T5 model~\cite{raffel2020exploring} as the foundational architecture, including T5-Small and T5-Base, which are grounded on encoder-decoder transformers~\cite{vaswani2017attention}.

\subsubsection{Results and Analysis} We report the average metrics of T5-Small and T5-Base merged from the two aforementioned tasks, as shown in Table~\ref{t5}. In NLP experiments, we utilize only 1\% of the training data for model evaluation in each episode, allowing for maximizing the merging speed, but undoubtedly posing a significant challenge to the robustness of RMM. Nevertheless, our method still demonstrates improvements of 9.97\% and 7.15\% compared to the baseline methods on two architectures. The experimental results prove the remarkable capabilities of RMM in NLP tasks, confirming its proficiency in addressing both similar and cross-domain model merging requirements. Moreover, RMM is also capable of merging varying quantities of model parameters and knowledge, which leads to forming a versatile language model.
\begin{table}[t]
  \centering
  \caption{\textbf{Ablation study and time consuming results.} Under different data ratios, we report the performance of RMM, the ablation of DAR, and the improvement in speed.}
  \label{dar}
  \begin{tabularx}{\columnwidth}{p{1.6cm}p{1.2cm}X<{\raggedleft}X<{\raggedleft}X<{\raggedleft}}
    \toprule
        \textbf{Metric}  
                         &\textbf{Method} 
                         &\textbf{100\%} 
                         &\textbf{10\%}
                         &\textbf{1\%}\\
    \midrule
        \multirow{2}{*}{Acc \#1}   & RMM     & 71.37  & 69.02   & 74.44 \\
                                   & w/o DAR & 71.37  & 69.52   & 71.45 \\
    \midrule
        \multirow{2}{*}{Acc \#2}   & RMM     & 76.35  & 78.58   & 72.79 \\
                                   & w/o DAR & 76.35  & 76.42   & 75.10 \\
    \midrule
        \multirow{2}{*}{Average}   & RMM     & 73.86  & \textbf{73.80}   & \textbf{73.61} \\
                                   & w/o DAR & 73.86  & 72.97   & 73.27 \\
    \midrule
    \midrule
        Time           &\multirow{2}{*}{RMM}  & 3.86h    & 0.37h  & 0.04h     \\
        Speed Up       &                      & 1.0×     & 10.4×  & 96.5× \\
    \bottomrule
  \end{tabularx}
\end{table}
\subsection{Additional Results and Analysis}
\subsubsection{Ablation Study} 
\begin{figure}[t]
    \centering
    \includegraphics[width=0.97\columnwidth]{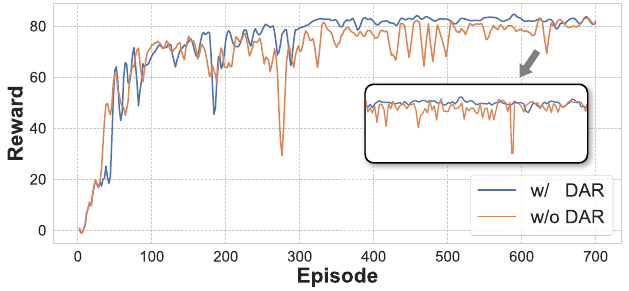}
    \caption{\textbf{Episode-Reward variation.} We illustrate the variation in reward per episode, indicating the advantages of DAR in enhancing search performance.} 
    \label{reward}
\end{figure}

To enhance the efficiency of the interaction between the agent and the environment in RMM, DAR was introduced to reduce the fraction of evaluation data and to dynamically adjust the rewards received by the agent. We conduct an ablation study on DAR using various data ratios and evaluate the accuracy of the ViT-B/16 architecture merged from CUB-200 and Stanford Dogs. Specifically, we tested three data ratios: 100\%, 10\%, and 1\%, then summarize the results in Table~\ref{dar}. It is evident that DAR can improve the performance of RMM in few data environments, which elevates the average accuracy of the merged model by 0.83\% and 0.34\%, respectively. Additionally, we illustrate the variations in reward per episode under 1\% data ratio in Fig.~\ref{reward}, where the adjustment of DAR stabilizes the overall reward feedback, leading to a more superior merged model.

\subsubsection{RMM Acceleration}
The proposed DAR has significantly enhanced the efficiency of RMM, enabling the rapid acquisition of the desired merged models. Notably, RMM eliminates the need for gradient calculations in the original models, as it solely focuses on connecting their distinct layers for evaluation inference. Such ability of conserving computational resources is particularly valuable, allowing users to quickly merge and deploy multi-task models on lightweight devices. In Table~\ref{dar}, we present the time consumption and speed up achieved using 10\% and 1\% of data compared to the traditional search-based settings with a full dataset. It can be seen that, even with merely 1\% of data, RMM can still merge models in 0.04 hours (i.e. less than 3 minutes) without a noticeable decline in performance, achieving an impressive speed up of approximately a hundred times. We hope our approach can provide new insights into edge model merging, offering a more comprehensive merging strategy across diverse scenarios.
\section{Conclusion}
In this paper, we presented RMM, a unified model merging framework that leverages reinforcement learning to integrate model parameters, distinguishing it from traditional training-free methods. RMM is both flexible and efficient, capable of handling various base models and merging algorithms, with acceleration up to a hundred times thanks to the DAR mechanism. Extensive experiments demonstrate that RMM achieves state-of-the-art performance across diverse tasks, effectively addressing the limitations of existing methods. Future work will focus on extending RMM to more complex scenarios, such as multi-modal and heterogeneous model merging, and exploring adaptive strategies to further enhance its versatility.

\section*{Acknowledgment}
This work is supported by Zhejiang Province High-Level Talents Special Support Program "Leading Talent of Technological Innovation of Ten-Thousands Talents Program" (No. 2022R52046).

\bibliographystyle{IEEEbib}
\bibliography{icme2025references}

\end{document}